\pdfoutput=1

\documentclass[11pt]{article}

\usepackage[]{emnlp2021}

\usepackage{times}
\usepackage{latexsym}
\usepackage{graphicx}

\usepackage[T1]{fontenc}

\usepackage[utf8]{inputenc}

\usepackage{microtype}

\newcommand{\figref}[1]{\hyperref[#1]{Fig.\ \ref*{#1}}}
\newcommand{\tabref}[1]{\hyperref[#1]{Table\ \ref*{#1}}}
\newcommand{\secref}[1]{\hyperref[#1]{Section\ \ref*{#1}}}
\newcommand{\algoref}[1]{\hyperref[#1]{Algorithm\ \ref*{#1}}}

\title{Azimuth: Systematic Error Analysis for Text Classification}

\author{Gabrielle Gauthier-Melançon, 
  Orlando Marquez Ayala, Lindsay Brin, Chris Tyler, \\ \textbf{Frédéric Branchaud-Charron, Joseph Marinier, Karine Grande, Di Le} \\
  ServiceNow \\
  \texttt{
\{gabrielle.gm,orlando.marquez,lindsay.brin\}@servicenow.com}}

\begin{document}
\maketitle
\begin{abstract}
We present Azimuth, an open-source and easy-to-use tool to perform error analysis for text classification. Compared to other stages of the ML development cycle, such as model training and hyper-parameter tuning, the process and tooling for the error analysis stage are less mature. However, this stage is critical for the development of reliable and trustworthy AI systems. To make error analysis more systematic, we propose an approach comprising dataset analysis and model quality assessment, which Azimuth facilitates. We aim to help AI practitioners discover and address areas where the model does not generalize by leveraging and integrating a range of ML techniques, such as saliency maps, similarity, uncertainty, and behavioral analyses, all in one tool. Our code and documentation are available at \url{github.com/servicenow/azimuth}.

\end{abstract}

\section{Introduction}

As academic and research labs push the boundaries of artificial intelligence, more and more enterprises are including NLP models\footnote{While models are part of pipelines that can include pre- and post-processors, we use the term models for simplicity.} in their real-world systems. This is exciting yet risky due to the complexity of current deep learning models and their increasing social impact. Whereas in traditional software development, engineers have methods to trace errors to code, it is challenging to isolate sources of error in AI systems.

NLP models can suffer from poor linguistic capabilities  \cite{ribeiro-etal-2020-beyond}, hallucinations \cite{https://doi.org/10.48550/arxiv.2202.03629}, learning spurious correlations via annotation artifacts \cite{gururangan-etal-2018-annotation}, or amplifying social biases \cite{chang-etal-2019-bias, DBLP:journals/corr/abs-2112-14168}. In addition to the adverse effects that these problems can cause to both individuals and society, deploying problematic models can lead to legal and financial penalties \cite{hbr.burt.new-ai-regulations.2021}. Even when poorly functioning models have no adverse social effects, they can still degrade trust, leading to problems with user adoption \cite{kocielnik-2019-accept-imperfect-ai,mckendrick.joe}.

\begin{figure}[t]
    \centering
    \includegraphics[width=\linewidth]{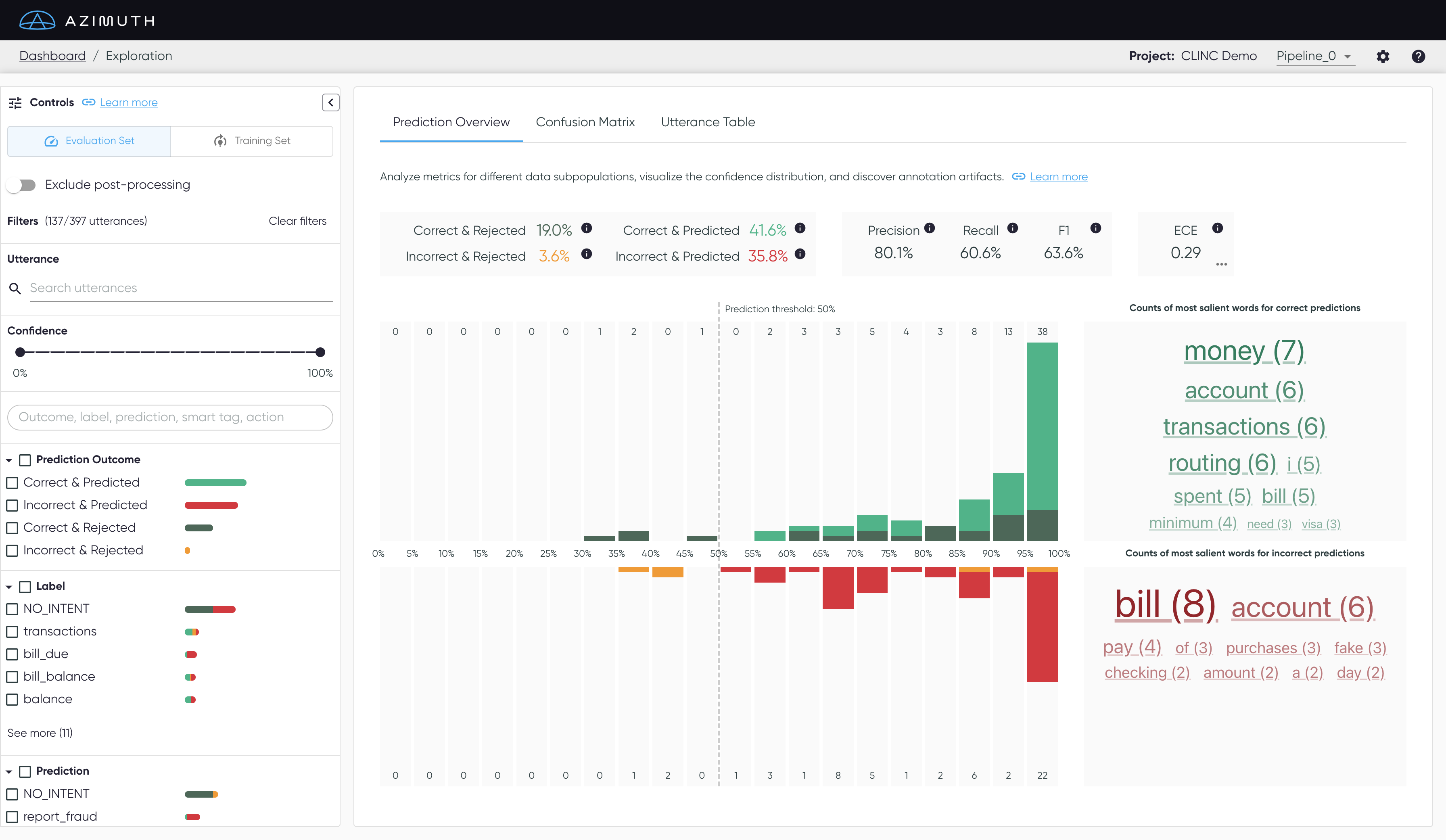}
    \caption{The exploration space of Azimuth allows users to explore dataset and pipeline quality across different data subpopulations.}
    \label{fig:azimuth_performance_overview}
\vspace*{-0.5cm}
\end{figure}

While it may be unreasonable to expect a perfect model, AI practitioners should communicate existing limitations so that stakeholders can make informed decisions about deployment, risk mitigation, and allocation of resources for further improvements \cite{arnold2019factsheets, mitchell2019model}. However, the current common error analysis practices may not be sufficiently thorough to provide this visibility, limiting our capacity to build safe, trustworthy AI systems.

As part of the ML development cycle, practitioners choose appropriate metrics based on business requirements. These metrics should assess whether NLP models have acquired the linguistic skills to perform the specified task. While metrics are useful to rapidly compare different models, they are only the beginning of quality assessment as they do not expose failure modes, i.e., types of input where the model fails, and do not readily indicate what can be improved. Furthermore, relying on metrics alone can be harmful, as they can overestimate quality and robustness while hiding unintended biases \cite{ribeiro-etal-2020-beyond, bowman-dahl-2021-will}.

To discover and address these limitations, error analysis is crucial. Unfortunately, compared to other stages of the ML development cycle, the error analysis stage is less mature. While various tools are commonly used to train and tune neural networks\footnote{Hugging Face Transformers, Ray Tune, etc.}, there is less convergence and adoption of both standards and tools to analyze errors. Analysis is typically conducted using custom scripts, spreadsheets, and Jupyter notebooks, guided by a practitioner's intuition, which may help uncover some problems while missing others. Large evaluation sets make per-example analysis of incorrect predictions time-consuming and tedious. Because this error analysis process is often informal, ad hoc, and cumbersome, practitioners often skip or rush this stage, focusing solely on high-level metrics. This hinders traceability and accountability, and introduces risks, especially for models deployed in production.

To alleviate these issues, we contribute the following:
\vspace{-3mm}
\begin{itemize}
    \itemsep-0.2em
    \item A systematic and intuitive error analysis process that practitioners can follow to improve their ML applications.
    \item Azimuth, an open-source and easy-to-use tool that facilitates this process, making thorough error analysis of text classification models easier and more accessible.
\end{itemize}

\section{Systematic Error Analysis}
\label{sec:systematic-error-analysis}

\begin{figure*}[h]
\caption{Proposed approach for systematic error analysis.}
\centering
\includegraphics[width=\textwidth]{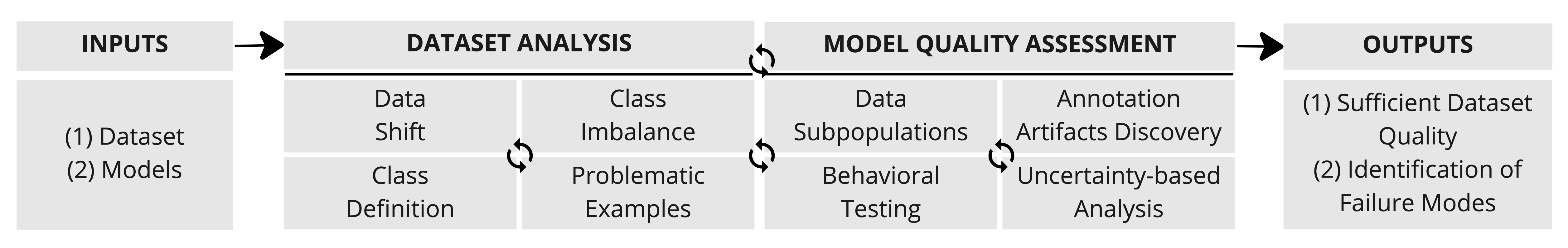}
\vspace*{-0.3cm}
\label{fig:process}
\end{figure*}

The goal of error analysis is to understand where and why a model succeeds or fails, to better inform both model improvement and deployment decisions. We propose a process grouped into two categories: (1) dataset analysis and (2) model quality assessment, as illustrated in \figref{fig:process}. Dataset analysis involves assessing and improving the data used to train and evaluate the model, while quality assessment focuses on model behavior. This process should be iterative, as analyzing model predictions can help identify and fix dataset problems, and dataset findings can help explain and reduce model errors. A limitation of our approach is that it does not currently assess whether models are learning a task ethically. While some of the techniques we refer to below can be used for this purpose, building ethical NLP systems is outside the scope of this paper.  

\subsection{Dataset Analysis}

Dataset analysis is indispensable to validate that the available data is appropriate for the given ML task and to suggest possible approaches to solve it. It should be conducted both before and throughout the ML development cycle, as it can guide model improvements. This analysis should encompass all data splits, although held-out evaluation sets require special care to avoid overfitting. High-quality data is essential as it impacts both model quality and the choice of models that will be deployed \cite{curtis-et-al.2021}. 

We identify four common types of dataset problems: data shift, class imbalance, class definition, and problematic examples. While these may not apply to all tasks, they illustrate common data problems. The first three can be detected by analyzing data at the dataset level, while problematic examples require example-level analysis. 

\textbf{Data Shift.} The training and validation splits must be compared to identify significant data shift caused by inadequate sampling or poor labeling practices (e.g., duplicated examples, missing classes). Taking into consideration the challenges in creating high-quality datasets \cite{bowman-dahl-2021-will}, one must guard against significant differences in data distributions across splits as they may cause quality problems or lead to poor choices in model selection and training. 

\textbf{Class Imbalance.} In classification tasks, class imbalance occurs when there are differences in the proportion or number of examples in each class relative to each other. This is problematic because models often overpredict classes that are overrepresented in the training data, whereas class imbalance in the evaluation data may cause performance metrics to be misleading. If unwanted class imbalance is present, practitioners may choose to resolve the issue through data augmentation for specific classes, or by upsampling or downsampling.  

\textbf{Class Definition.}  The classes used in the ML task affect what the model will learn. Practitioners may need the assistance of domain experts to determine whether the classes are defined appropriately. For instance, class overlap is a symptom of poor class definition that can be seen when groups of examples in multiple classes overlap semantically, resulting in classes that are not easily separable. Some semantic overlap may be acceptable and not cause model confusion, whereas other overlap may indicate low data quality or poor dataset construction, leading to poor model performance. Improving class definition can include relabeling mislabeled examples, better defining classes by merging or splitting them, and augmenting data for specific classes. 

\textbf{Problematic Examples.} Multiple kinds of challenges can arise at the example level. This includes examples outside the expected domain, “difficult” examples with exceptional characteristics (e.g., words from other languages, long sentences) but for which model performance is desired, examples for which humans would disagree on the label, and accidental mislabeling issues. These problems can confuse class boundaries and cause model errors. Resolution options include relabeling examples, removing examples from datasets, and targeted data augmentation. 
\vspace{-.1cm}
\subsection{Model Quality Assessment}
\vspace{-.05cm}
Once datasets are deemed sufficient in quality, practitioners train and tune models aiming to find the best candidates. Quality assessment consists of examining how well the selected models perform on a specified evaluation set as well as their ability to generalize beyond the evaluation set. The objective is to understand where and why the model fails. Issues exposed in model quality assessment can also inspire further dataset improvements.

Assessing quality by examining metrics is the aspect of error analysis that is typically performed, as it is relatively fast to conduct \cite{church2019survey}. This quantitative assessment allows for quick comparison across models or model versions using known scores such as precision, recall, or F1. This can include metrics to measure model calibration, such as expected calibration error (ECE), as well as metrics that quantify business value, safety, and bias. 

As practitioners already tend to analyze high-level metrics to have a basic notion of model quality, we focus here on other approaches to evaluate generalization. To help discover and correct failure modes before models are deployed, we propose four types of evaluation: (1) assess model quality according to data subpopulations, (2) discover annotation artifacts, (3) perform behavioral testing, and (4) conduct uncertainty-based analysis. 

\textbf{Data Subpopulations} are subsets of datasets with shared characteristics such as examples with long text, entities, keywords, same label, etc. Models may behave differently on these subpopulations, but high-level metrics can hide these failure modes. Guided by domain knowledge, this analysis can help ensure that the model generalizes well under various input characteristics. Often, performance on problematic subpopulations can be improved by targeted data augmentation.

\textbf{Annotation Artifacts} are patterns in the labeled data that can be exploited by models to learn simple heuristics instead of learning to perform the task, and yet achieve high metric scores \cite{gururangan-etal-2018-annotation}. The result is a model that relies undesirably on specific features, such as words that tend to correlate with the label only in the available datasets. Annotation artifacts can be discovered by leveraging feature-based explainability methods, such as saliency maps, to approximately determine the importance of every token when the model makes a prediction. Once such artifacts are found, practitioners can refine the datasets to help models decrease their reliance on them.

\textbf{Behavioral Testing} assesses model robustness by validating model behavior based on input and output \cite{ribeiro-etal-2020-beyond}. Perturbing a dataset and observing the corresponding predictions can help identify important errors, biases, or other potentially harmful aspects of the model that may not be otherwise obvious. For production models, this type of testing is critical, as the range of user input is infinite, while the evaluation datasets are finite. Models that change predictions or their confidence values when the input is slightly altered without changing its semantics can have unintended consequences and may lose the user's trust. 

\textbf{Uncertainty-based Analysis} includes assessing examples that are more difficult for the model to learn. Examination of lower-confidence predictions can help indicate regions of the data distribution where the model may fail in the future, similar to Data Maps \cite{swayamdipta-etal-2020-dataset}. A more sophisticated approach is to find predictions with high epistemic uncertainty, computed with techniques such as Bayesian deep ensembles \citep{wilson2020bayesian}. Augmenting the training dataset with more representative examples from these data regions can improve model quality.

\vspace{3mm}
Systematic error analysis will help practitioners obtain the following:
\vspace{-2mm}
\begin{itemize}
\itemsep0em
    \item Datasets deemed sufficient in quality that can be used for training and evaluation. 
    \item Identified failure modes: known situations where the model fails to generalize.
\end{itemize}
\vspace{-2mm}
To help practitioners achieve these outcomes, we contribute Azimuth to the NLP community. 
\vspace{.1cm}
\section{Azimuth, an Open-Source Tool}
\vspace{.1cm}

Azimuth was developed as an internal tool at ServiceNow and open sourced in April 2022. It facilitates our proposed approach to systematize error analysis of ML systems. While it is currently tailored for text classification, it could be extended to support other use cases. The tool was built by a cross-collaborative team of scientists, engineers, and designers, with a human-centered approach that focuses on the AI practitioner's needs. 

Before launching Azimuth, the user defines in a configuration file the dataset splits and pipelines to load and analyze. In Azimuth, pipelines refer to the ML model as well as any pre-processing and post-processing steps. The tool is built on top of the Hugging Face (HF) datasets library \cite{lhoest-etal-2021-datasets} and easily interfaces with HF pipelines \citep{Wolf_Transformers_State-of-the-Art_Natural_2020}. For flexibility, any Python function can be defined as a pipeline. If pipelines are unavailable, Azimuth can still be used, although with limited features, by reading predictions from a file. The tool can also be used for dataset analysis without a pipeline.

\subsection{User Workflow}
\begin{figure}[h]
\caption{On the dashboard, pipeline quality is broken down by different data subpopulations.}
\centering
\includegraphics[width=\linewidth]{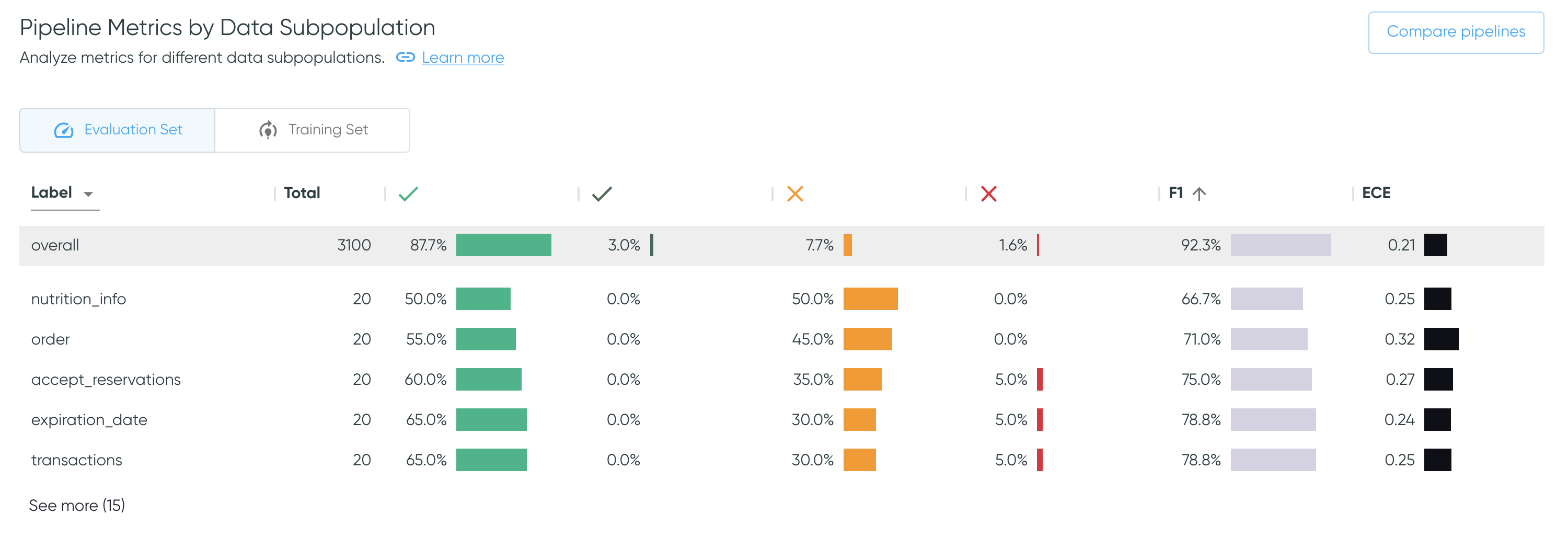}
\label{fig:model_quality}
\vspace{-0.8cm}
\end{figure}

Azimuth leads users through two main screens: the dashboard and the exploration space. At startup, users are brought to the dashboard, which presents a summary of the different capabilities and flags potential dataset and pipeline problems (\figref{fig:dashboard}). The dashboard is linked to the exploration space through pre-selected filters that allow a more detailed investigation of issues raised on the dashboard. For example, as shown in \figref{fig:model_quality}, pipeline quality is indicated by metrics on different data subpopulations, such as label, prediction, and smart tag (defined below). By clicking on a row, the user is brought to the exploration space filtered on the corresponding subpopulation.

\begin{figure}[h]
\caption{The exploration space displays examples along with their predictions and smart tags. The user can propose actions to improve the dataset.}
\centering
\includegraphics[width=\linewidth]{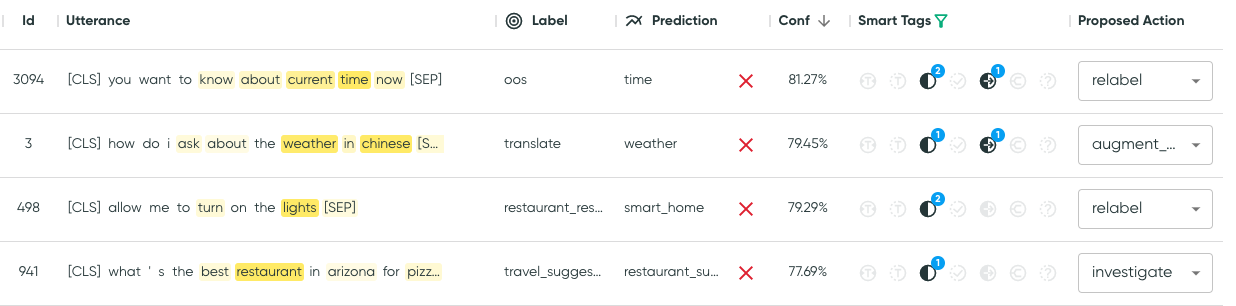}
\label{fig:utterances}
\vspace{-1cm}
\end{figure}

The exploration space allows the user to explore the dataset and pipeline quality by filtering on different data subpopulations, model confidence, and/or the presence of certain text. Data can be shown for predictions both before and after post-processing (e.g., thresholding). Changing the filters will update all available visualizations and tables split over three tabs. The first tab displays several quality metrics, a histogram of confidence scores for correct and incorrect predictions, and word clouds indicating word importance (\figref{fig:azimuth_performance_overview}). The second tab displays the confusion matrix (\figref{fig:guidance_colors}). The last tab shows the details of the raw data and predictions, allowing users to inspect individual examples and their smart tags, and propose actions for those that are problematic (\figref{fig:utterances}). Clicking on a particular example takes users to its details page, which provides additional information such as the prediction at each post-processing step, behavioral test results, and semantically similar examples from each dataset split (\figref{fig:example_travel_time}). 

\vspace{.1cm}
\subsection{Smart Tags}
\vspace{.1cm}
An integral feature of Azimuth is the concept of smart tags, which are tags assigned to dataset examples based on predefined rules. Smart tags allow users to explore the dataset and model predictions through various data subpopulations, helping to detect failure modes and to identify problematic examples. Some smart tags are straightforward (examples with few tokens), while others are more complex (examples with high epistemic uncertainty). They are grouped in families, based on their associated capabilities, which may require access to only the dataset or to both the dataset and the pipeline. 

To assist the process of improving the dataset, Azimuth has a “proposed action” field with options to indicate further actions that should be taken on specific examples. There are natural correspondences between certain smart tags and proposed actions. Users can filter examples in the exploration space by a certain smart tag to focus specifically on an aspect of dataset analysis. For instance, the high epistemic uncertainty smart tag can detect hard-to-predict examples that may require actions such as relabeling, removing examples, or augmenting the training set with similar examples. In the same vein, smart tags from the similarity analysis may highlight examples that suggest the need to add new classes or merge existing ones. There is also a generic proposed action \emph{investigate} to signal that further troubleshooting is required as no concrete action is clear yet.

\vspace{.1cm}
\subsection{Capabilities}
\begin{figure*}[h]
\caption{The steps of our proposed approach are pictured in gray boxes while the corresponding Azimuth capabilities are in black boxes. "All smart tags" means that all capabilities and their smart tags can be useful for this step.}
\centering
\includegraphics[width=\textwidth]{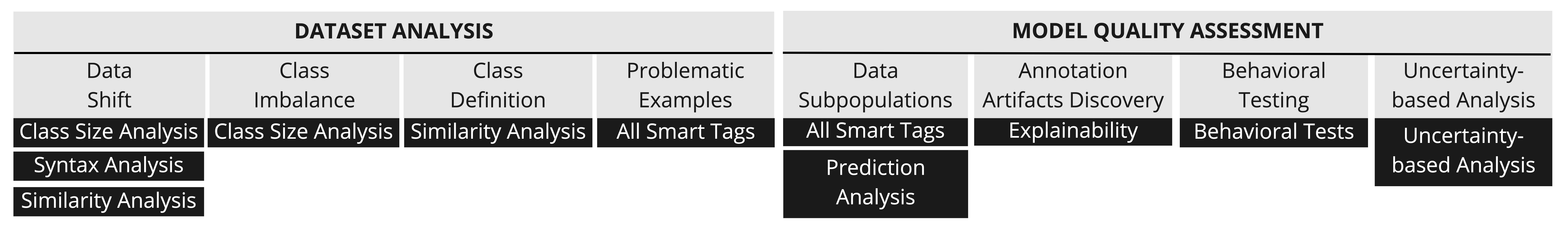}
\label{fig:capabilities}
\vspace{-1cm}
\end{figure*}

\vspace{.1cm}
Azimuth provides a variety of capabilities for both dataset analysis and model quality assessment. \figref{fig:capabilities} depicts how the different capabilities map to the approach proposed in section \ref{sec:systematic-error-analysis}. Our documentation includes a detailed list of each feature and smart tag that are part of Azimuth's capabilities.   

\textbf{Class Size Analysis.} Azimuth surfaces classes having too few examples, which may indicate class imbalance within a split. The tool also detects a form of dataset shift by raising warnings when the number of examples across classes is not similarly distributed across splits.

\textbf{Syntax Analysis.} The Spacy library \cite{Honnibal_spaCy_Industrial-strength_Natural_2020} is used to inspect the syntax of examples. Besides detecting significant differences in sentence length between dataset splits, we use dependency trees and part-of-speech tags to explore model behavior when examples are missing a subject, verb, or object. Syntactic smart tags can help identify problematic examples and explore model performance on atypical syntax. 

\textbf{Similarity Analysis.} Azimuth leverages the SentenceTransformers framework \cite{reimers-2019-sentence-bert} to compute sentence embeddings. We use these embeddings to calculate the similarity of all pairs of examples in all dataset splits, and perform similarity search using faiss \cite{johnson2019billion}. Smart tags and nearest neighbors based on similarity values allow for surfacing potential dataset shift, class overlap, or problematic examples. For example, some smart tags surface examples that are distant from their nearest neighbors, while others identify examples where the neighbors belong to a different class.

\textbf{Prediction Analysis.} Visualizations and metrics, such as the confusion matrix and the expected calibration error, help assess model quality. This type of analysis also includes threshold comparison as well as smart tags that compare prediction results across different pipelines.

\textbf{Explainability.} We generate saliency maps \cite{kindermans2019reliability} to reveal why a particular prediction was made in terms of the relevance of each input token to the prediction \cite{bastings-filippova-2020-elephant, atanasova-etal-2020-diagnostic}. We use a gradient-based technique as it is fast to compute and available for all token-based models.

In the exploration space, saliency maps are shown for each example in the filtered subpopulation. Additionally, word clouds allow inspection of the most salient words for correct (green) and incorrect (red) predictions, which can help identify words present in examples that the model struggles to classify. Differences between word clouds for a specific data subpopulation, such as class label, can hint at spurious correlations or a model's over-reliance on specific words, as exemplified in section \ref{subsubsec:annotation_artifacts}. Lastly, on the example details page, the user can compare the example's most salient words to those of similar training examples, which may help explain misclassification.

\textbf{Behavioral Analysis.} Azimuth uses behavioral testing to help assess the general linguistic capabilities of NLP models. As an initial implementation, we use NLPAug \cite{ma2019nlpaug} and custom functions to create Robustness Invariance tests: input perturbations that should not change the model predictions \cite{ribeiro-etal-2020-beyond}. Predictions obtained with and without perturbations are compared to reveal areas where the model lacks robustness. Smart tags highlight examples whose predictions have changed unexpectedly, helping to identify specific problems. In addition, users can define new functions to test other linguistic capabilities, such as those proposed in \textsc{CheckList}. 

\textbf{Uncertainty-based Analysis.} Users can explore examples based on model confidence and visualize their confidence distribution in a histogram. Some smart tags highlight predictions that are almost correct, based on model confidence. Filtering out these predictions can help focus on examples that are more problematic for the model, which often hint at mislabeling or larger dataset issues, such as poorly defined labels. In contrast, focusing on the almost correct predictions can help identify class overlap or issues that may be addressed by targeted data augmentation. For models with dropout, we additionally use Baal \cite{atighehchian2019baal} to compute a smart tag that surfaces examples with high epistemic uncertainty, which have a greater chance of being problematic.

\subsection{Design}
UX design and research are essential to create valuable and usable ML systems. Our design priorities for Azimuth were focused on supporting the error analysis process by increasing efficiency, balancing guidance with flexibility and user control, and fostering user delight. Through collaborative design sessions, workshops, and user interviews with AI practitioners, we improved Azimuth quickly and iteratively. In particular, user interviews uncovered several challenges that we addressed with design modifications, including disentangling different levels of analysis and preventing choice paralysis (details in \ref{sec:design_challenges}). 

As a result, Azimuth's design approach generally follows a paradigm of guided exploration, which shaped the creation of features such as the dashboard and the control panel on the exploration space. Additionally, the navigation and progressive disclosure lead users to discover important features and take action quickly by separating, but linking, high-level warnings and detailed investigation. We make the process enjoyable and efficient by including visualizations and the ability to search, filter, hide and show information as needed. The content and communications provide context and guidance without being obstructive. For example, Azimuth prioritizes contextual information icons or subtle explanations, and our color system helps to assign priority and call attention to warnings and errors. See Appendix \ref{sec:app_design} for details.

\subsection{Extensibility}
To customize the error analysis experience, users can easily change a variety of settings in Azimuth's configuration file. For instance, users can change the encoder used for the similarity analysis, or the thresholds that determine class imbalance.

Azimuth capabilities are implemented via \texttt{Modules}, which use a dataset, a configuration, and optionally a model to perform the desired analyses with distributed computing and caching. Our repository contains details on how to add a new \texttt{Module}\footnote{\url{https://github.com/ServiceNow/azimuth/tree/main/azimuth/modules}}.



For more complex modifications, we encourage the community to submit issues on our GitHub page. As the process of error analysis continues to be refined, we hope that Azimuth will grow along with the community.

\subsection{Case Study}
\vspace{-.1cm}
We verified the utility of our methodology and Azimuth by applying them to a DistilBert model trained on CLINC-OOS~\citep{larson-etal-2019-evaluation}. This large intent classification dataset has 150 “in-scope” classes spanning several domains and one Out-of-Scope (OOS) class. Our goal is to demonstrate how Azimuth’s features can efficiently direct users to specific, resolvable problems, even with CLINC-OOS's large size and wide topic range. Below we summarize the most salient findings while Appendix \ref{sec:app_case_study} includes more details. 

\vspace{-2mm}
\begin{itemize}
    \itemsep-0.2em
    \item \textit{no\_close} tags (26 examples) revealed classes with discordant semantic spaces across dataset splits.
    \item \textit{conflicting\_neighbors} smart tags surfaced overlapping class pairs. For some examples, multiple labels were applicable, possibly warranting a multi-label classification model.
    \item Despite the relatively clean nature of the dataset, \textit{conflicting\_neighbors} smart tags surfaced 25 mislabeled examples in the training set and 18 in the validation set. 
    \item Accuracy was worse than average on several data subpopulations, including short sentences ($\sim$8\% worse than long sentences), examples lacking a verb ($\sim$10\% worse than average), and examples failing the punctuation robustness test ($\sim$20\% worse than average).
    \item Word clouds revealed possible annotation artifacts such as model dependence on a specific verb or the plural form of a particular noun.
    \item Behavioral testing showed a high failure rate for typos ($\sim$25\%), surfacing classes for which the model depended on specific tokens. 
    \item The model is underconfident, warranting temperature scaling or a lower threshold. 
\end{itemize}
\vspace{-2mm}

Overall, our analysis revealed multiple issues that could be addressed through data cleaning and augmentation. Moreover, the high overall validation accuracy (94\%) hides a lack of robustness.

\section{Related Work} 

There exist other tools that can help practitioners evaluate their NLP systems beyond observing metrics. Broadly speaking, these solutions are implemented as component libraries, standalone applications, or some combination thereof. 

Small components have a lower investment of effort to get started but may require more technical expertise to use. While they can produce results quickly, they fail to address the problem of ad-hoc processes and may lead to a "paradox of choice" \cite{goel-etal-2021-robustness}. Examples include the \textsc{CheckList} Python package \cite{ribeiro-etal-2020-beyond} that can be used for behavioral testing and the AllenNLP Interpret toolkit \cite{wallace-etal-2019-allennlp} that computes gradient-based saliency maps.

On the other hand, standalone applications may require more effort to set up and it may not be obvious how to integrate them into the ML development process. Their benefits are that, once setup, less technical practitioners can use them and their usage can be standardized. CrossCheck \cite{arendt-etal-2021-crosscheck} and Robustness Gym \cite{goel-etal-2021-robustness} are used as Jupyter widgets, while Errudite \cite{wu-etal-2019-errudite} and Language Interpretability Tool (LIT) \cite{tenney-etal-2020-language} exist both as standalone applications and as widgets in Jupyter notebooks. 

Currently, Azimuth is a standalone application that requires low setup effort and can be integrated into the ML development cycle as proposed in section \ref{sec:systematic-error-analysis}. At the same time, we provide the benefits of existing third-party component libraries, added as features in an easy-to-use interface. We built Azimuth to assist in performing comprehensive error analysis using a single tool by including functionality found elsewhere: filtering and analysis of behavior on subpopulations (CrossCheck, Errudite, LIT, Robustness Gym), input variations such as counterfactual error analysis, robustness testing (Errudite, LIT, Robustness Gym), model comparison (CrossCheck, LIT), and explainability techniques (LIT, AllenNLP Interpret).

\section{Conclusion} 
We propose a systematic approach to error analysis as an iterative workflow between dataset analysis and model quality assessment. We contribute Azimuth to the NLP community in order to facilitate this approach. Future work includes expanding our capabilities, exposing potential ethical concerns in the data or ML models, and extending Azimuth to cover other tasks and domains. We welcome open-source contributions.

\section*{Ethical Considerations}
We believe our proposed approach to systematize the error analysis stage of the ML development cycle should help decrease the adverse social effects that error-prone models can create as well as increase user adoption and trust. An important gap in our approach is that it does not explicitly suggest techniques that can detect whether models behave ethically or whether datasets used to train these models contain harmful biases.

\section*{Acknowledgements}
We thank ServiceNow for sponsoring the development of Azimuth, especially our colleagues involved in building and publicizing it. We also thank ServiceNow teams and users in the NLP community for their useful feedback.

\bibliography{anthology,custom}
\bibliographystyle{acl_natbib}
\clearpage

\appendix

\section{Azimuth Design}
\label{sec:app_design}

The Azimuth team employed human-centered design methods to create interfaces that propose a workflow to AI practitioners while taking into account the exploratory nature of their work. 

\subsection{Design Challenges}\label{sec:design_challenges}

User interviews surfaced functional challenges that we were able to address with design modifications. We present several examples below.

\begin{figure}[h]
\caption{The dashboard was added to address the need to separate but link different levels of analysis.}
\centering
\includegraphics[width=\linewidth]{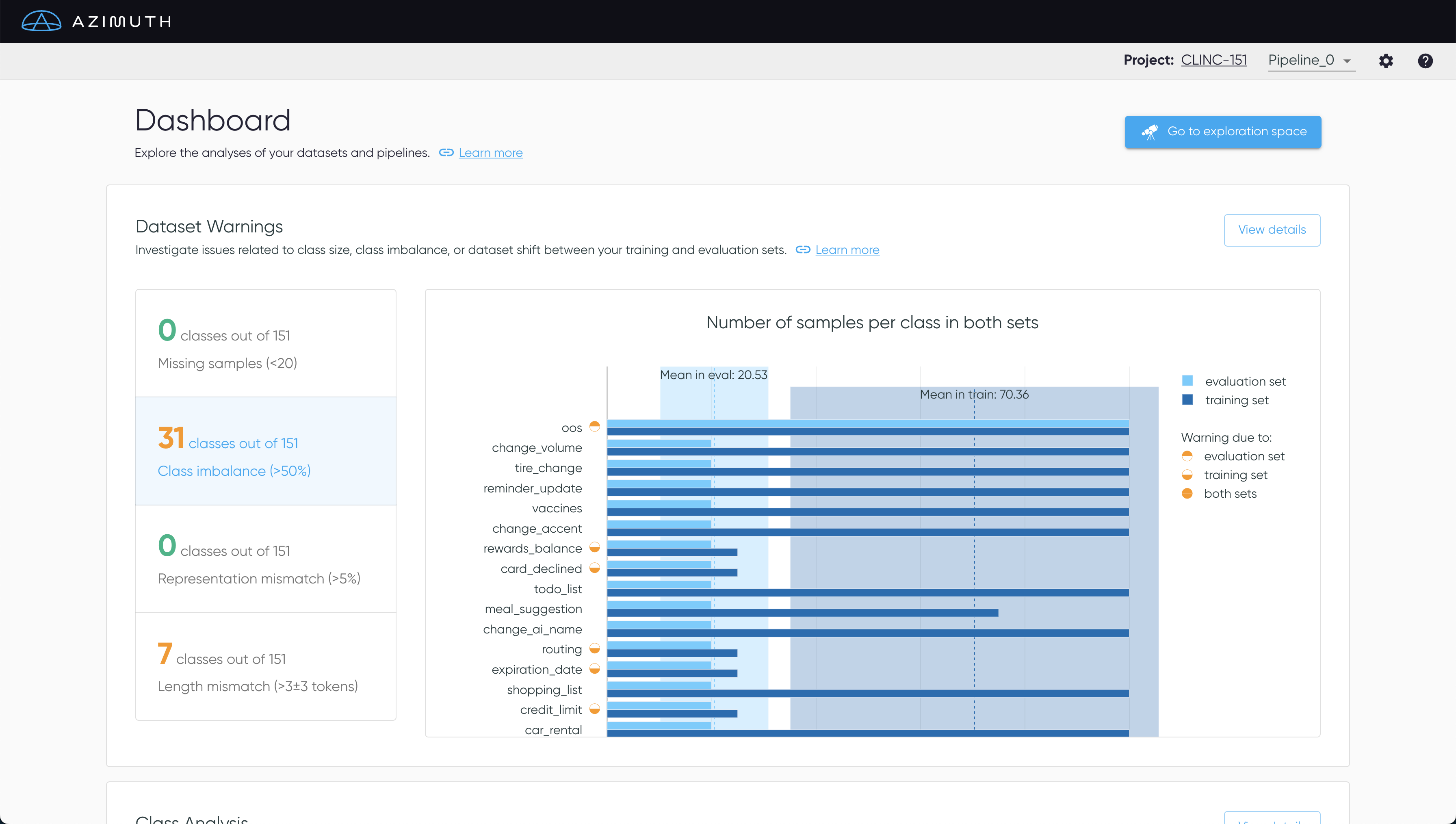}
\label{fig:dashboard}
\end{figure}

First, early design iterations centered the user experience around the exploration space, allowing for unstructured analysis and exploration of individual examples. However, this space did not provide enough high-level insights nor sufficient guidance to inform a user's next actions. We created the dashboard (\figref{fig:dashboard}) to summarize these insights and guide the user's exploration of the dataset. 

Second, the notion of data subpopulations was included very early on in the form of filtering via individual dropdown menus. This did not show all possible filtering options nor their distributions across the dataset. As \figref{fig:usability_filtering} illustrates, the control panel that replaced the dropdown menus now shows users how filters can be combined as well as the filters that result in the largest number of model errors. For example, filtering the dataset by prediction \textit{oil\_change\_how} will result in a data subpopulation where the model performed very well (the line is almost entirely green).

Third, we grappled with the trade-off between providing users with all possible avenues of exploration versus presenting a limited set of options that are relevant in the context of the user's activity. We ended up replacing the initial option for users to select data subsets by individual confidence histogram bins, which led to choice paralysis or haphazardness, with the option to select data subsets by picking a confidence threshold. For example, users can now view all examples where the model assigned a confidence less than 80\%.

\subsection{Guidance}

\begin{figure}[h]
\caption{Example of Azimuth color system.}
\centering
\includegraphics[width=\linewidth]{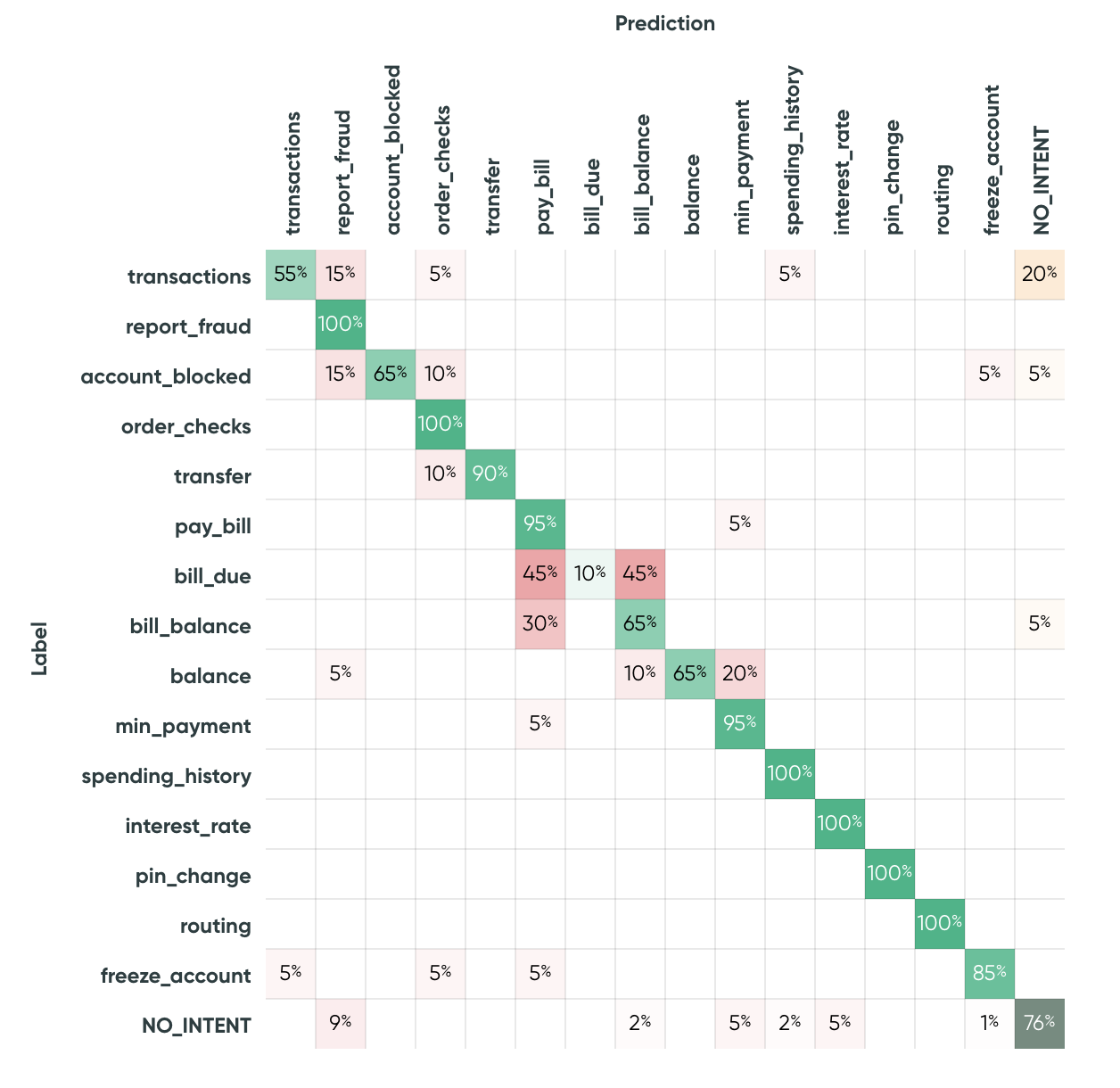}
\label{fig:guidance_colors}
\end{figure}

 \begin{flushleft}
\textbf{Colors.} A consistent color scheme is used to draw attention to warnings, as shown in the confusion matrix in \figref{fig:guidance_colors}.
\end{flushleft}

\begin{figure}[h]
\caption{Links to product documentation are embedded throughout the application.}
\centering
\includegraphics[width=0.45\textwidth]{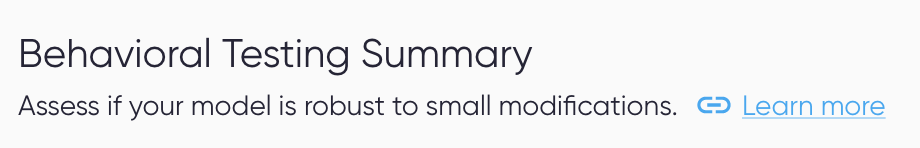}
\label{fig:guidance_doc}
\end{figure}

 \begin{flushleft}
\textbf{Documentation.} To help guide our users, each section element has direct links to our detailed product documentation, as shown in \figref{fig:guidance_doc}.
\end{flushleft}

\begin{figure}[h]
\caption{Tooltips help give context to users.}
\centering
\includegraphics[width=0.35\textwidth]{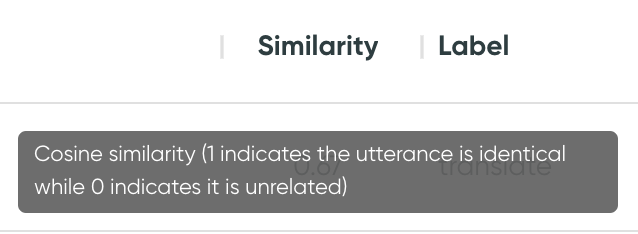}
\label{fig:guidance_tooltips}
\end{figure}

 \begin{flushleft}
\textbf{Tooltips.} Tooltips are also available on essential elements to identify and define key concepts, show calculation methods, define terms, and provide information on how to use certain functionality, as shown in \figref{fig:guidance_tooltips}.
\end{flushleft}

\subsection{Flexibility}

\begin{figure}[h]
\caption{Control panel provides sophisticated filtering.}
\centering
\includegraphics[width=0.35\textwidth]{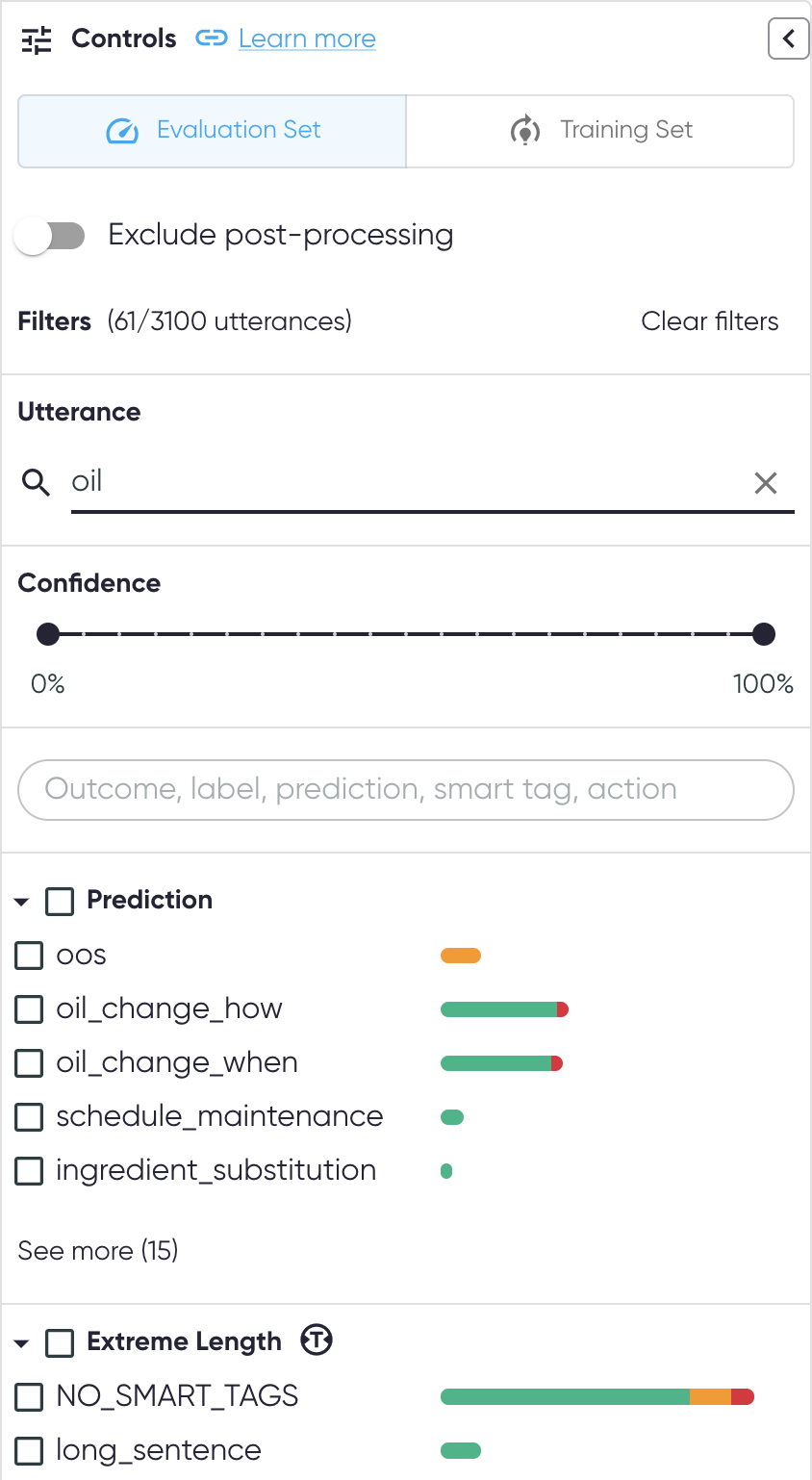}
\label{fig:usability_filtering}
\end{figure}

\textbf{Filtering.} A control panel, illustrated in \figref{fig:usability_filtering}, allows users to filter the dataset and predictions across diverse criteria, such as containing a particular word or having a certain confidence score. The user can also filter by label, predicted class, and all available smart tags. Predictions can also be shown with and without post-processing, which can help distinguish model errors from errors related to the post-processing steps.

\begin{figure}[h]
\caption{Users can sort by column and customize what they see.}
\centering
\includegraphics[width=0.35\textwidth]{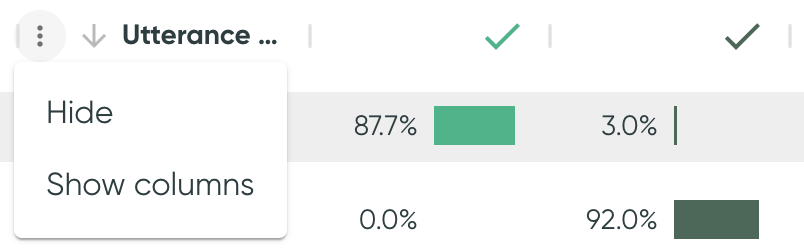}
\label{fig:usabilty_tables}
\end{figure}

 \begin{flushleft}
\textbf{Sorting and Column Customization.} As shown in \figref{fig:usabilty_tables}, many of our tables allow users to show/hide columns as they see fit, in addition to sorting the table information based on the content in the columns.
\end{flushleft}

\section{Case Study Details}
\label{sec:app_case_study}

\subsection{Dataset and Model Details}

We chose a DistilBert model \cite{DBLP:distilbert} from among the best ranking models on PapersWithCode\footnote{\url{https://paperswithcode.com/sota/text-classification-on-clinc-oos}}. The CLINC-OOS dataset and the DistilBert model were both downloaded from the Hugging Face Hub. We configured Azimuth to use a threshold of 0.5. When the model score was below this value, examples were classified as the rejection class, also known as Out-of-Scope (OOS). Azimuth provides the option to conduct analyses before or after post-processing (in this case, thresholding to OOS); we took advantage of this option for some analyses below. We used the \textbf{Imbalanced} training split and the validation split as the evaluation set.

\subsection{Dataset Analysis}

\begin{figure*}[ht]
\caption{Example of possible dataset shift on the example details page.}
\centering
\includegraphics[width=\textwidth]{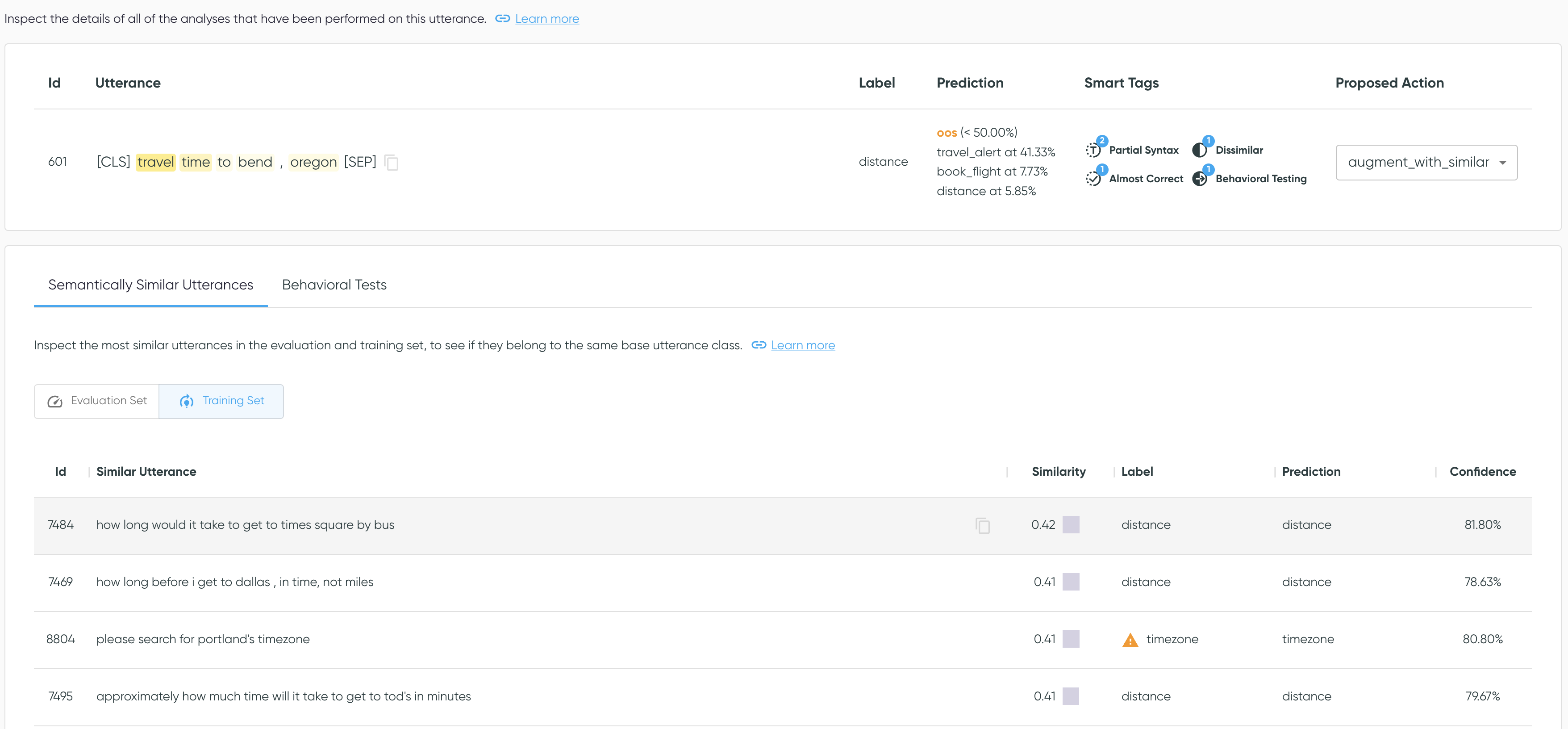}
\label{fig:example_travel_time}
\end{figure*}

\subsubsection{Dataset Shift}
We examined data shift by inspecting misclassified examples from the evaluation set using the \textit{no\_close\_train} smart tag, which identifies those having no training examples that are similar to them. This approach identified 26 examples that were candidates for targeted data augmentation. Data augmentation could be guided by looking at the example's most salient tokens and its most similar examples in the training set.

For instance, as shown in \figref{fig:example_travel_time}, an example mentioning "travel time" was labeled “distance”, but predicted as “travel alert” with low confidence (41\%). According to the saliency map, the model focuses on the words “travel” and “time”. Most similar examples in the training data were labeled as "distance" but did not contain the word “travel”. In addition, the examples in the training data containing the word “travel” were predominantly labeled “travel alert,” followed by “travel notification” and “vaccines”. Augmenting the class “distance” in the training data with examples containing words related to "travel time" could be a way to address this data shift issue.

\subsubsection{Class Imbalance}

Azimuth detects class imbalance in the training set, which is normal given the split that we chose. Another observation is that the OOS class is overrepresented in the evaluation set, compared to other classes.

\subsubsection{Class Definition}

The dataset covers a large variety of topics with its 150 intents, not all of which are at the same hierarchical level of semantic meaning. This can cause the model to have difficulty differentiating between intents. The \textit{conflicting\_neighbors} smart tags surfaced many examples for which it was not easy to determine the correct label, or where multiple labels could apply. This helped direct us to overlapping class pairs, including those that are effectively supersets of other intents. For example:

\begin{itemize}
\itemsep-0.2em
\item There is some ambiguity between the intents “restaurant suggestion” and “travel suggestion”. When the example refers to a restaurant in a particular city, the ground-truth label is "travel suggestion" rather than "restaurant suggestion", but the model did not learn this distinction.  
\item Some intents cover large semantic spaces, such as “translate” and “define”, making them difficult to predict correctly. For example, “how do I ask about the weather in chinese” is predicted as “weather” instead of "translate" because the text can be interpreted as a weather-related question.
\end{itemize}

\subsubsection{Problematic Examples}
\begin{figure}[h]
\caption{Subset of problematic examples.}
\centering
\includegraphics[width=0.45\textwidth]{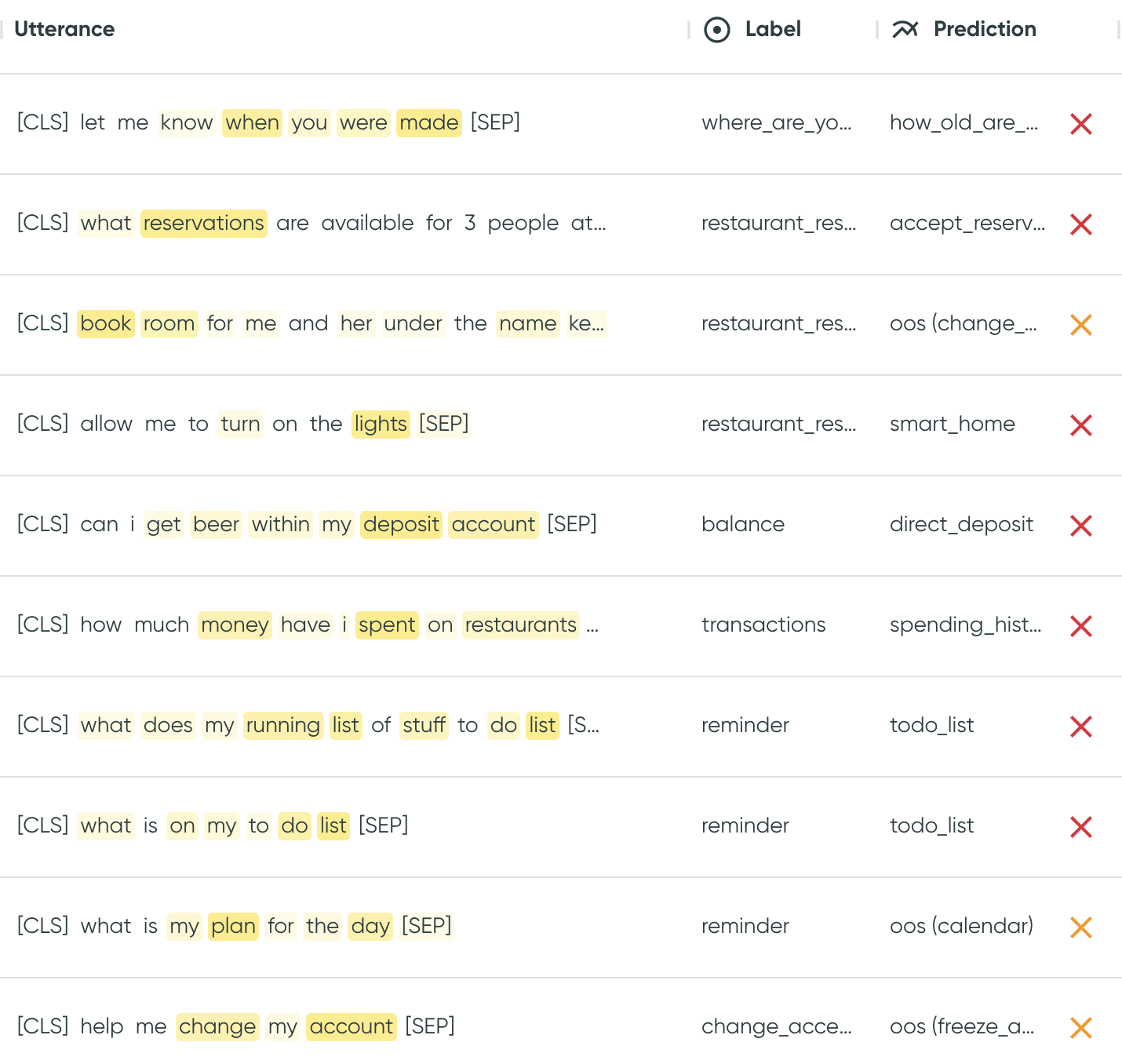}
\label{fig:mislabeled}
\end{figure}

Using \textit{similarity} smart tags (\textit{conflicting\_neighbors} and \textit{no\_close}), we were able to detect 18 mislabeled examples in the evaluation set and 25 in the training set. Notably, we also found a mislabeling pattern where four examples were labeled “change accent” instead of “change account”. This dataset is relatively clean and, for better or worse, does not reflect the messiness and ambiguity of real-world data. Although these examples make up a small proportion of the dataset, it is notable that we were able to surface them relatively easily. \figref{fig:mislabeled} shows a subset of the mislabeled examples. 

\subsection{Model Quality Assessment}

The model is 99.2\% accurate on the training set and 93.9\% accurate on the evaluation set without post-processing. The errors in the evaluation set are either misclassifications (5.8\%) or rejections to OOS (0.3\%). When adding a threshold of 0.5, the accuracy decreases to 90.7\%. On the other hand, misclassification errors decrease to 1.6\% and OOS gets predicted more often instead of predicting the wrong in-scope class (7.7\%). 

\subsubsection{Data Subpopulations}
We examined misclassification rates for different data subpopulations, such as label and a variety of smart tags. Several interesting issues were quickly revealed: 

\begin{itemize}
\itemsep-0.2em
\item The model has lower accuracy on a few intents when evaluated on the training set, especially when examples contain the word "name", which can be found in intents such as "what is your name" and "change user name" (respectively $\sim$75\% and $\sim$82\% accuracy on the training set). These observations on the training data already tell us that some intents are more difficult to learn than others.

\item Compared to average evaluation set accuracy, the model performs better (+3\%) on long sentences (more than 15 tokens) and worse (-5\%) on short sentences (less than three tokens). Short sentences are more often predicted as OOS. The model also performs worse on examples having no verbs (often short sentences), with a drop of $\sim$12\% in accuracy compared to the average.

\item As expected, the model has lower accuracy on examples with conflicting or few similar examples. Compared to average, we see a drop in accuracy of $\sim$25\% on examples tagged with the \textit{conflicting\_neighbors} smart tag and a drop of $\sim$10\% on examples tagged with the \textit{no\_close} smart tags.  

\item The accuracy is lower than average ($\sim$20\%) on examples that fail any punctuation robustness tests (\textit{failed\_punctuation} smart tag). These tests are considered to fail if the prediction changes when the punctuation is altered. 

\end{itemize}

\subsubsection{Annotation Artifacts Discovery}\label{subsubsec:annotation_artifacts}
\begin{figure}[h]
\caption{Important words are shown for correct and incorrect predictions for the "accept reservations" intent.}
\centering
\includegraphics[width=0.45\textwidth]{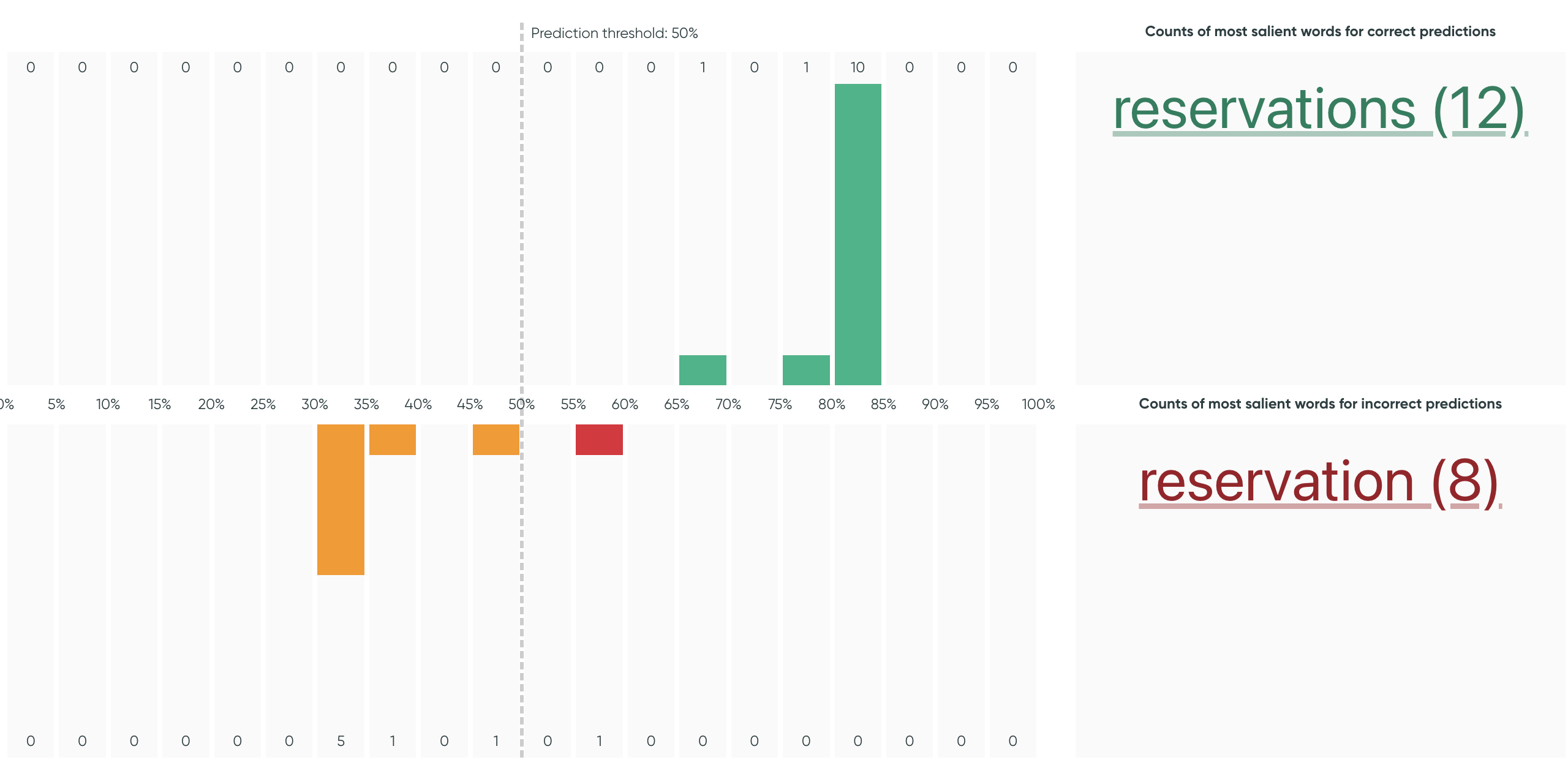}
\label{fig:ann_art}
\end{figure}

Exploration with Azimuth revealed the following cases where model predictions depended on specific words or word forms, as well as other cases described below in “Behavioral testing”. As  
shown in \figref{fig:ann_art}, predicting "accept reservations" intent is highly dependent on the word “reservations”. When examples in the evaluation set contain the singular form "reservation", the model makes mistakes.

\subsubsection{Behavioral Testing}\label{subsubsec:checklist}

Azimuth flagged the high failure rate of behavioral tests on both the training and evaluation sets, due largely to the high failure rate when introducing typos (approximately 22-25\% failure across several tests for both dataset splits). Additionally, model predictions changed for some examples when altering the punctuation, especially when introducing a comma or a period. Although the overall failure rate for this test was low ($\sim$2\%), users may consider this type of failure unacceptable. Together, these tests suggest a robustness issue that should be addressed through dataset augmentation or pipeline design.  

Further exploration of individual examples, via saliency maps and smart tags for behavioral tests, revealed several intents that were highly dependent on single tokens, indicating potential annotation artifacts. For instance, the model fails when “luggage” is misspelled in examples labeled as “lost luggage”. Similarly, it fails when “rewards” is misspelled in examples labeled as “redeem rewards”.

\subsubsection{Uncertainty-based Analysis}
The model is generally underconfident, having a maximum confidence of 90\% and an expected calibration error (ECE) of 0.21 on the evaluation set. This could be addressed via temperature scaling. The model is particularly underconfident on specific intents, such as "what is your name" and "pto used", with top confidence scores around 60\% to 70\%.  

The \textit{high\_epistemic\_uncertainty} smart tag surfaced 48 examples, most of them labeled as OOS. The model only has an accuracy of $\sim$30\% on these examples, demonstrating how difficult they are to classify. Some examples include "idk", labeled as "maybe", and "what's today's high and low", labeled as "weather".

\end{document}